\documentclass[letterpaper, 10 pt, conference]{ieeeconf}
\IEEEoverridecommandlockouts
\usepackage{amsmath,amssymb,amsfonts}
\usepackage{algorithmic}
\usepackage{graphicx}
\usepackage{textcomp}
\usepackage{xcolor}
\usepackage{lipsum}
\usepackage{multirow}
\usepackage{booktabs}
\usepackage{array}
\usepackage{url}
\usepackage{cite}
\title{\LARGE \bf Depth Edge Alignment Loss: DEALing with Depth in Weakly Supervised Semantic Segmentation}
\author{Patrick Schmidt$^{1}$, Vasileios Belagiannis$^{2}$ and Lazaros Nalpantidis$^{1}$ 
\thanks{This work has been funded and supported by the EU Horizon Europe project ``RobetArme” under the Grant Agreement 101058731. We extend our thanks to the DTU Computing Center for providing computational resources.}
\thanks{$^{1}$DTU - Technical University of Denmark, Kongens Lyngby, Denmark
       {\tt\small \{pasch, lanalpa\}@dtu.dk}}
\thanks{$^{2}$Friedrich-Alexander-Universität Erlangen-Nürnberg, Germany
       {\tt\small vasileios.belagiannis@fau.de}}
}
\begin{document}
\maketitle
\begin{abstract}
Autonomous robotic systems applied to new domains require an abundance of expensive, pixel-level dense labels to train robust semantic segmentation models under full supervision. This study proposes a model-agnostic Depth Edge Alignment Loss to improve Weakly Supervised Semantic Segmentation models across different datasets. The methodology generates pixel-level semantic labels from image-level supervision, avoiding expensive annotation processes. While weak supervision is widely explored in traditional computer vision, our approach adds supervision with pixel-level depth information, a modality commonly available in robotic systems. We demonstrate how our approach improves segmentation performance across datasets and models, but can also be combined with other losses for even better performance, with improvements up to $+5.439$, $+1.274$ and $+16.416$ points in mean Intersection over Union on the PASCAL VOC / MS COCO validation, and the HOPE static onboarding split, respectively.
Our code is made publicly available\footnote[3]{\url{https://github.com/DTU-PAS/DEAL}}.
\end{abstract}
\section{Introduction}
Autonomous robotic systems rely on robust perception capabilities to ensure smooth and safe operation without human intervention. A key component to this is semantic segmentation for precise localization of objects in a scene. These models are usually trained fully-supervised with a plethora of images and pixel-level, dense labels which are expensive to obtain, and for new applications often not readily available. Weakly Supervised Semantic Segmentation (WSSS) addresses this by obtaining pixel-level dense labels from weaker forms of supervision, such as Class Activation Maps (CAM) extracted from image-level classification labels \cite{ahn_learning_2018, wang_self-supervised_2020,li_uncertainty_2021, chen_self-supervised_2022, lee_threshold_2022, chen_class_2022}. While WSSS shows promising results on general-purpose datasets like PASCAL VOC \cite{everingham_pascal_2010} or MS COCO \cite{coco_dataset}, it has not been widely explored in a robotics context yet.
We argue that WSSS can be extended to operate in real-world robotics scenarios, by incorporating additional information such as depth---commonly available to robots due to the omnipresence of RGB-D sensors.
We therefore propose a new, model-agnostic Depth Edge Alignment Loss (DEAL), which enforces edges extracted from a CAM to align with edges extracted from the depth map of the current scene. 
DEAL is added on top of other losses, and can be combined with other improving losses like Importance Sampling Loss (ISL) and Feature Similarity Loss (FSL) \cite{jonnarth_high-fidelity_2024}, boosting those even further.
\par
\begin{figure}[t]
    \centering
    \includegraphics[width=0.8\linewidth]{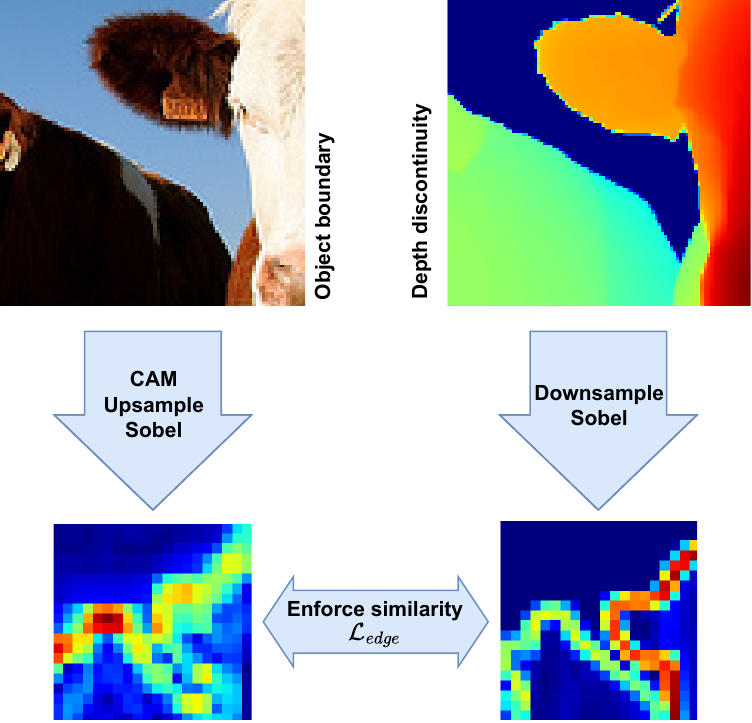}
    \caption{A graphical summary of the intuition behind our proposed Depth Edge Alignment Loss. Alignment of object boundaries in RGB images and edges extracted at depth discontinuities can improve the results of WSSS.}
    \label{fig:approach_summary}
\end{figure}
We observe that including depth information in the training process of existing WSSS methods can improve the segmentation performance in terms of mean Intersection over Union (mIoU) from weak supervision. Depth provides pixel-level dense information about the spatial layout of a scene and---as opposed to pixel-level semantic labels---is inexpensive to obtain through RGB-D sensors commonly used in autonomous robotic systems. In Multi-Task Learning, where multiple down-stream tasks share one common feature extraction backbone and thus a common feature representation, research has shown high task affinity between semantic segmentation and monocular depth estimation, provided sufficient network capacity \cite{standley_which_2020}. Our added loss shows that it can leverage this affinity, as it improves mIoU performance across datasets and models. In addition, even when using real-life noisy RGB-D data, our proposed DEAL shows improvements, showing it is robust to noise. 
While our approach prerequisites CAM-based WSSS models, which is the most-explored mode of generating dense pixel-level labels from image-level supervision, it can inspire recent WSSS approaches using Vision Language Models to include depth information in their training and inference processes. Our contributions can be summarized as follows:
\begin{itemize}
    \item We introduce our novel Depth Edge Alignment Loss (DEAL) which improves mIoU segmentation performance in CAM-based WSSS frameworks.
    \item We demonstrate how DEAL can improve segmentation performance \textit{across datasets and models}.
    \item We show that our depth-guided WSSS can yield strong improvements for robotic applications, even in the presence of noisy depth data.
\end{itemize}
\section{Related Work}
\subsection{Weakly Supervised Semantic Segmentation (WSSS)}
WSSS has been proposed as an alternative to Fully Supervised Semantic Segmentation (FSSS). FSSS requires pixel-level semantic labels, which are expensive to obtain.  In contrast, in the most common case, WSSS uses much easier to obtain image-level category labels for training. Other forms of weak supervision in the WSSS task include scribbles, bounding box, or foreground/background point annotations \cite{akiva_single_2023}.  To highlight the added time requirements to obtain pixel-level labels, on the MS COCO dataset, it took workers about 85,000 hours to obtain these for 328,000 images \cite{coco_dataset}, including 20,000 hours for image-level labeling and 10,000 hours for foreground point annotations. Commonly, image-level labels are used to train an image classification network, then generate a coarse localization of objects using Class Activation Mappings (CAM) which are further refined using dense Conditional Random Fields (CRF) \cite{krahenbuhl_efficient_2011} and finally used to train a segmentation network on the extracted, pixel-level pseudo labels \cite{ahn_learning_2018, wang_self-supervised_2020,li_uncertainty_2021, chen_self-supervised_2022, lee_threshold_2022, chen_class_2022}. 
As outlined, WSSS usually builds upon a multi-stage framework of classification training, pseudo-label generation and refinement, and segmentation training. Single-stage, end-to-end approaches do exist \cite{pinheiro_image-level_2015, papandreou_weakly-_2015, tang_regularized_2018, zhang_reliability_2019, araslanov_single-stage_2020, akiva_single_2023}, but they generally do not match the performance of multi-stage approaches. 
In our experiments, we focus on approaches using image-level classification labels and on the first two stages of the framework, classification network training and CAM generation. 
\par
Recent approaches leverage foundation models like the Segment Anything (SAM) model \cite{sun_alternative_2023, chen_segment_2023, jiang_segment_2023, yang_foundation_2023}, however, since SAM was trained with mask supervision where human mask refinement was involved \cite{Kirillov_2023_ICCV}, we argue that these frameworks are not WSSS approaches, given the influence of pixel-level labels through SAM. 
With the rise of Vision Transformer (ViT)-based architectures, some works leverage the self-localizing properties of this architecture \cite{xu_multi-class_2022, zhu_weaktr_2023, xu_mctformer_2024}. The similarity to the Transformer architecture commonly used in Natural Language Processing (NLP) enabled the development of Vision-Language models like Contrastive Language-Image Pretraining (CLIP) \cite{radford_learning_2021}. In a WSSS context, recent approaches leveraged CLIP to generate initial pseudo-mask seeds, departing from CAM-based approaches \cite{deng_qa-clims_2023, xu_learning_2023, lin_clip_2023, lin_semantic_2025, zhang_frozen_2025, zhu_weakclip_2025, jang_dial_2025, yang_exploring_2025, xu2025toward}. In our work, we demonstrate our method both on a ViT-based and a ResNet-based WSSS framework, showcasing its model-agnostic property.
\subsection{WSSS on non-standard datasets}
Typically, studies on WSSS present their approaches on two standard, general-purpose datasets: PASCAL VOC and MS COCO \cite{ahn_learning_2018, wang_self-supervised_2020,li_uncertainty_2021, chen_self-supervised_2022, lee_threshold_2022, chen_class_2022}. These datasets have certain characteristics that do not necessarily hold true in a real-life robotics scenario: they contain close to no object co-occurrence, a small number of classes per image and a large number of negative examples per class \cite{kim2024weakly}. Opposed to the standard WSSS studies, \cite{kim2024weakly} shows a CLIP-based approach on driving scenes from the CityScapes dataset under high object co-occurrence ratios. Akiva and Dana \cite{akiva_single_2023} present a single-stage WSSS approach and analyzes other datasets like CityScapes or Cranberry from Aerial Imagery (CRAID) in regard to object diversity, scale and count. However, their method relies on point annotations as opposed to image-level classification labels, providing an initial localization cue. Chan et al. \cite{chan_comprehensive_2021} demonstrate a performance gap of WSSS approaches on histopathology and satellite images.  
In addition to the standard benchmarking sets for WSSS, we conduct experiments on HOPE \cite{tyree_6-dof_2022}, a dataset for robotic manipulation of household objects. 
\subsection{Depth Integration in Weakly Supervised Settings}
Most WSSS approaches solely rely on RGB images as input modality. While the text modality is getting more common with CLIP-based WSSS approaches, the text used in these approaches is just a transformed version of the image-level classification label. When approaching WSSS from a robotic---instead of a pure computer vision---perspective, modalities from common onboard sensors like RGB-D arise. Previous research in Multi-Task Learning has shown high task affinity between semantic segmentation and monocular depth estimation \cite{standley_which_2020}. To the best of our knowledge, \cite{ergul_depth_2022} is the only WSSS approach including depth data in the training process, where depth is used as an additional modality in the CRF refinement of CAMs. While being an unsupervised semantic segmentation approach, \cite{Sick_2024_CVPR} includes a depth correlation mechanism to correlate segmentation features with their spatial distance, pushing feature vectors of pixels with big spatial distance apart and pulling them together if the spatial distance is low. Opposed to \cite{ergul_depth_2022}, we employ our approach before CAM refinement, avoiding expensive CRF calculations during training. We also follow a different intuition than \cite{Sick_2024_CVPR} and get inspired by the observation that depth boundaries often co-occur with semantic boundaries. 
\section{Method}
\begin{figure*}[htbp]
    \centering
    \includegraphics[width=0.95\linewidth]{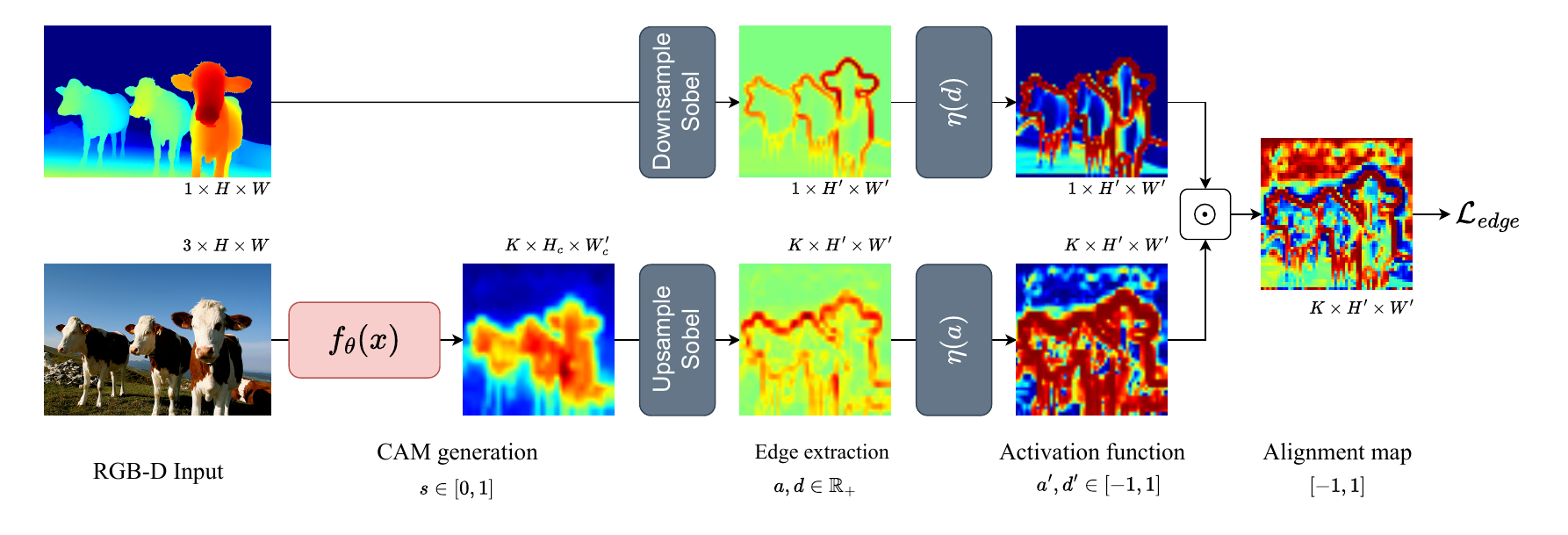}
    \caption{Detailed graphical overview of our method. The upper strand shows the depth information pipeline, and the lower strand shows the CAM generation pipeline. From both CAM and depth, we extract edges $a$ and $d$, use a $\tanh$ activation function $\eta$, calculate the alignment map through channel-wise per-element multiplication and then aggregate those into $\mathcal{L}_{\mathrm{deal}}$. Note that the only trainable module is $f_\theta(x)$, which can be any trainable CAM-based WSSS framework.}
    \label{fig:approach_detailed}
\end{figure*}

We present our DEAL which acts as an additional loss function for any CAM-based WSSS approach. Figure \ref{fig:approach_detailed} shows a high-level overview of our proposed method. Assuming the availability of a RGB-D dataset $\mathcal{D} = \{x_i,y_i,D_i\}_{i=1}^N$ with images $x$, image-level multi-label classification labels $y$ and depth maps $D$, we adopt CAM-based WSSS approaches with a trainable model $f_\theta$. Our method adds a loss function $\mathcal{L}_{\mathrm{edge}}$ to the training processes of each approach to encourage alignment of edges extracted from CAMs $s$ with edges extracted from depth maps $D$ to improve mIoU segmentation performance. Finally, we combine $\mathcal{L}_{\mathrm{edge}}$ with other losses to further boost performance.
\subsection{CAM Generation}
We deal with the problem of inferring class-wise binary per-pixel foreground/background masks $M\in \{0,1\}^{K\times H\times W}$ , i.e., we use a model $f_\theta(x) $ with input images $x \in \mathbb{R}^{3\times H\times W}$ which outputs CAMs $s \in \mathbb{R}^{K\times H_c \times W_c}$. We adopt ViT-based and ResNet-based models as $f_\theta$ to generate $s$: WeakTr \cite{zhu_weaktr_2023} with an ImageNet pretrained DeiT-S backbone and SEAM \cite{wang_self-supervised_2020} with a ImageNet pretrained ResNet38 backbone. The first two columns of the bottom row in Fig. \ref{fig:approach_detailed} illustrate the process of obtaining $s$. We train the model $f_\theta$ using image-level labels $y \in \{0,1\}^{K}$ only, e.g. by using a multi-label soft margin loss $\mathcal{L}_{\mathrm{MLSM}}(\hat{y}, y)$, with $\hat{y} = \mathrm{Agg}(s) \in \mathbb{R}^K$ being the predicted class probabilities from the CAM output  using the aggregation function $\mathrm{Agg}: \mathbb{R}^{K \times H_c \times W_c} \rightarrow \mathbb{R}^{K}$.
In both WeakTr and SEAM, $\mathrm{Agg}$ is a global average pooling (GAP) layer.
\subsection{Depth Edge Alignment Loss}
We propose DEAL, an additional loss intuitively aligning extracted edges of CAMs $s$ with extracted edges of depth maps $D\in \mathbb{R}^{1\times H\times W}$. The loss is added to the training losses of WeakTr and SEAM, $\mathcal{L}_{\mathrm{WeakTr}}$ and $\mathcal{L}_{\mathrm{SEAM}}$ respectively. Our method performs the following steps to calculate DEAL, which we denote $\mathcal{L}_{\mathrm{edge}}$: 
\begin{enumerate}
    \item Extract edges from CAMs $s$ and depth maps $D$
    \item Apply an activation function $\eta(\cdot)$ to the extracted edges $a$ and $d$ respectively 
    \item Perform a pixel-wise multiplication of the edge activation map
    \item Aggregate those into a single score $\mathcal{L}_{\mathrm{edge}}\in\mathbb{R}$. 
\end{enumerate}
First, we downsample the depth map from its original resolution $D \in \mathbb{
R}^{1\times H\times W}$ to a lower resolution $D' \in \mathbb{R}^{1\times H' \times W'}$ using bicubic interpolation:
\begin{equation}
    D' = \mathrm{Downsample}(D)
\end{equation}
If the generated CAMs $s$ with resolution $H_c\times W_c$ do not match this resolution, we upsample those to $s' \in \mathbb{R}^{K\times H' \times W'}$, using bicubic interpolation too:
\begin{equation}
    s' = \mathrm{Upsample}(s)
\end{equation}
We then extract the edges of both $s'$ and $D'$ using a $3\times3$ Sobel filter, and extract the magnitude: 
\begin{equation}
    \begin{aligned}
    \mathbf{G}_x = \begin{bmatrix}
        -1 & 0 & +1 \\
        -2 & 0 & +2 \\
        -1 & 0 & +1
    \end{bmatrix}
    \hspace{2em}
    \mathbf{G}_y = \begin{bmatrix}
        -1 & -2 & -1 \\
         0 &  0 &  0 \\
        +1 & +2 & +1
    \end{bmatrix} 
    \end{aligned}
    \label{eq:sobel}
\end{equation}
\begin{minipage}{0.45\linewidth}
\begin{equation}
    \begin{aligned}
    s'_{x} &= \mathbf{G}_x * s' \\
    s'_{y} &= \mathbf{G}_y * s' \\
    a &= \sqrt{{s'_x}^2 + {s'_y}^2} \\
    a &\in \mathbb{R_+}^{K\times H'\times W'}
    \end{aligned}
    \label{eq:cam_edges}
\end{equation}
\end{minipage}
\hfill
\begin{minipage}{0.45\linewidth}
\begin{equation}
    \begin{aligned}
    D'_{x} &= \mathbf{G}_x * D' \\
    D'_{y} &= \mathbf{G}_y * D' \\
    d &= \sqrt{{D'_x}^2 + {D'_y}^2}\\
    d &\in \mathbb{R_+}^{1 \times H' \times W'}
    \end{aligned}
    \label{eq:depth_edges}
\end{equation}
\end{minipage}\\
$\mathbf{G}_x$ and $\mathbf{G}_y$ in Eq. \ref{eq:sobel} are the convolutional kernels to extract vertical and horizontal edges respectively. In Eqs. \ref{eq:cam_edges} and \ref{eq:depth_edges}, $*$ denotes the convolution operation. Note that we perform Eqs. \ref{eq:cam_edges} for each class individually. Given that non-edge regions with magnitudes close to zero do not produce any gradient for the model parameters $\theta$ and we want DEAL to discourage CAM boundaries in regions with no depth edges, we apply an activation function $\eta(\cdot)$ to both $a$ and $d$.
Therefore,  we draw inspiration from \cite{jonnarth_high-fidelity_2024} and apply $\tanh$ activation to a log-transformed version of both $a$ and $d$ in Eqs. \ref{eq:activation_a} and \ref{eq:activation_d} respectively. Following \cite{jonnarth_high-fidelity_2024}, we use $\mu=2.5$.
\begin{equation}
        a' = \eta(a) = \tanh\left(\mu+{\log\left(\frac{a}{1-{a}}\right)}\right)\\
        \label{eq:activation_a}
\end{equation}
\begin{equation}
    d' = \eta(d) = \tanh\left(\mu+\log\left(\frac{d}{1-d}\right)\right)
    \label{eq:activation_d}
\end{equation}
This results in CAM edge and depth edge activations $a' \in [-1,1]^{K\times H'\times W'}$ and $d' \in [-1,1]^{1\times H'\times W'}$, with non-edge regions being mapped to -1 and strong magnitudes mapped to 1. The spatial activation maps are now aggregated into a single score $\mathcal{L}_{\mathrm{edge}}$---we perform a per-class, pixel-wise multiplication as in Eq. \ref{eq:deal}.
\begin{equation}
    \begin{aligned}
    \mathcal{L}_{\mathrm{edge}} &= -\frac{1}{HW}\sum_{ij=1}^{HW}\frac{1}{\sum_{c=1}^Ky_k}\sum_{k=1}^{K}y_k a'_{k,ij}d'_{ij}\\
    \end{aligned}
    \label{eq:deal}
\end{equation}
For each pixel $ij$, we calculate a per-class mean over $K$, multiplying the activation values with the ground truth label $y \in [0,1]^K$, where $y_k=1$ if class $k$ is present in the current sample and $y_k=0$ otherwise. Therefore, we ignore classes not present in the current sample. Finally, we take the average over pixels. In Eq. \ref{eq:deal}, aligned regions exert positive values and unaligned regions negative values. Therefore, since we minimize the loss, we multiply the loss with -1 to obtain $\mathcal{L}_{\mathrm{edge}}$. Figure \ref{fig:approach_detailed} summarizes the methodology and provides visualizations of the intermediate outputs.
\subsection{Auxiliary Losses}
We combine DEAL with ISL/FSL presented in \cite{jonnarth_high-fidelity_2024} to further boost mIoU performance. Note that this is existing work, and we outline the approach for a complete presentation of our framework.
ISL is an added classification loss using image-level labels $y$ and predictions, $\hat{y}'$, e.g. a multi-label soft margin loss $\mathcal{L}_{\mathrm{MLSM}}(\hat{y}, y)$. However, ISL uses Importance Sampling (IS) instead of global average pooling as the aggregation function $\mathrm{Agg}$ to obtain the predicted class probabilities $\hat{y}'$ from CAMs $s$:
\begin{equation}
    \hat{y}' = \mathrm{Agg}(s) = \mathrm{IS}(s)
\end{equation}
IS samples pixel locations $i_k,j_k$ using the normalized CAM $s_k \in [0,1]^{H_c\times W_c}$ as sampling probabilities for each class $k$ to obtain the prediction vector $\hat{y}'$, i.e. we draw $K$ pixels and concatenate the scores $y_k = s_{k,i_k,j_k}$ into $\hat{y}'$. Finally, we calculate the classification loss, e.g. $\mathcal{L}_{\mathrm{MLSM}}(\hat{y}',y)$, repeat this process $N$ times and take the average:
\begin{equation}
    \mathcal{L}_{\mathrm{is}} = \frac{1}{N}\sum_{n=1}^N\mathcal{L}_{\mathrm{MLSM}}(\hat{y
    }'_{n}, y)
\end{equation}
We follow \cite{jonnarth_high-fidelity_2024} and set $N=10$.
FSL is an added loss, encouraging high prediction distances $s_{ij},s_{uv}$ for dissimilarly colored pixel pairs $x_{ij},x_{uv}$ and vice versa. Given the CAM $s=f_\theta(x)$ for an RGB image $x$, $\mathcal{L}_{\mathrm{fs}}$ is calculated as follows:
\begin{equation}
    \begin{aligned}
        \mathcal{L}_{\mathrm{fs}}(x,s) &= -\frac{1}{HW}\sum_{ij,uv=1}^{HW}w_{ij,uv}g(s_{ij},s_{uv})\eta(\delta(x_{ij},x_{uv})) \\
    g(s_{ij}, s_{uv}) &= \frac{1}{2} \| s_{ij} - s_{uv} \|_2^2\\
    \delta(x_{ij},x_{uv}) &= \frac{1}{3}\|x_{ij} - x_{uv}\|_2
    \label{eq:distance_function}
    \end{aligned}
\end{equation}
with $w_{ij,uv}$ being the Gaussian spatial weight limiting the effect of this loss to the neighborhood of the current pixel considered, formulated as follows:
\begin{equation}
    w_{ij,uv} = \frac{1}{2\pi\sigma^2} \exp\left(-\frac{\left\|\begin{bmatrix}
        i\\
        j
    \end{bmatrix} - \begin{bmatrix}
        u\\
        v
    \end{bmatrix}\right\|_2^2}{2\sigma^2} \right) 
    \label{eq:gaussian_kernel}
\end{equation}
We follow \cite{jonnarth_high-fidelity_2024} and set $\sigma=5$. In contrast to \cite{jonnarth_high-fidelity_2024}, we use the L2 distance for $\delta$ instead of the L1 distance to leverage efficient GPU computations, observing minimal differences in training results.
\section{Experiments}
\subsection{Datasets, Metrics and Models}
We perform experiments on three datasets: PASCAL VOC 2012, MS COCO 2014 and HOPE. For PASCAL VOC, we additionally use data from the Semantic Boundaries Dataset (SBD) \cite{hariharan_semantic_2011} while training, but evaluate on the original training data. Furthermore, we demonstrate results on both WeakTr and SEAM, the first being ViT-based and the latter being ResNet-based to verify the applicability across architectures. We evaluate the methods on the train and validation splits, respectively. On HOPE, we use the dynamic and static onboarding sequences as training and validation data, respectively. Note that evaluation is done with ground-truth segmentation labels, which are not used during training, thus the train split is also indicative for validation purposes. We report the mean Intersection over Union (mIoU) over all classes, and report the mean over four different runs with different seeds. As in \cite{zhu_weaktr_2023}, for fair comparison, we use the checkpoint with the best segmentation mIoU on the training/validation set to generate CAMs, and use the best-performing threshold for each run when reporting metrics. Since PASCAL VOC and MS COCO do not contain ground-truth depth maps $D$, we generate those using a monocular depth estimator, Depth-Anything-V2-Base (DA-v2b) \cite{depthanything, depth_anything_v2}. For HOPE, we choose the dynamic onboarding sequences for training, since they include RGB-D data from an Intel RealSense camera. We do not preprocess the depth maps in any way.
\subsection{Baselines}
We adopt the default WeakTr configuration with a pretrained DeiT-S backbone and establish baseline results from training WeakTr over 60 epochs. Furthermore, we adopt ISL/FSL and apply it to WeakTr to analyze whether ISL/FSL boosts segmentation performance of WeakTr too, as this experiment has not been conducted in \cite{jonnarth_high-fidelity_2024}. For ISL/FSL, we apply each loss twice on WeakTr --- on the fine CAM branch and the coarse CAM branch. Finally, we aggregate them by taking the average. Therefore, we optimize the following:
 \begin{equation}
    \begin{aligned}
    \mathcal{L}_{\mathrm{is}} &= \frac{1}{2}\left(\mathcal{L}_{\mathrm{is, fine}} + \mathcal{L}_{\mathrm{is,coarse}}\right) \\
     \mathcal{L}_{\mathrm{fs}} &= \frac{1}{2}\left(\mathcal{L}_{\mathrm{fs, fine}} + \mathcal{L}_{\mathrm{fs,coarse}}\right) \\
     \mathcal{L}_{\mathrm{total}} &= (1 - w_{is})\mathcal{L}_\mathrm{WeakTr} + w_{is}\mathcal{L}_\mathrm{is} + w_{fs}\mathcal{L}_\mathrm{fs} .
     \end{aligned}
 \end{equation}
We scale both $\mathcal{L}_\mathrm{fs}$ and $\mathcal{L}_\mathrm{is}$ down by a factor of 0.1, and set $w_{is}=0.2,w_{fs}=1.0$, following training settings for MCTformer in \cite{jonnarth_high-fidelity_2024}---given the similarity in architecture to WeakTr. 
The baseline results are listed in Table \ref{tab:baseline_results_maxmIoU} for WeakTr training without ISL/FSL, with ISL/FSL, and with DEAL. A positive influence of ISL/FSL on mIoU segmentation performance can be observed with WeakTr too, validating the effect of the method on this model. 
\begin{table}[t]
    \caption{\normalfont{ Segmentation performance of CAMs generated by WeakTr trained on PASCAL VOC and MS COCO. We report the mIoU of each variant as a mean of four runs with four different seeds.}}
    \centering
    \small
    \setlength{\tabcolsep}{2pt}
    \begin{tabular}{@{}lll>{\centering\arraybackslash}p{0.15\linewidth}>{\centering\arraybackslash}p{0.15\linewidth}@{}}
    \toprule
    \multirow{2}{4em}{\textbf{Model}} & \multirow{2}{4em}{\textbf{Dataset}}& \multirow{2}{4em}{\textbf{Variant}}&
    \multicolumn{2}{c}{\textbf{mIoU} $\uparrow$}  \\
    &&&train&val\\
    \midrule
    \multirow{8}{4em}{\textbf{WeakTr}} & \multirow{4}{4em}{VOC}&Baseline&63.461 & 60.635\\
    && DEAL & 64.792 & 61.908 \\
    && ISL/FSL & 66.545 & 63.818 \\
    && DEAL + ISL/FSL & \textbf{66.876} & \textbf{64.671}\\
    \cmidrule(lr){2-5}
    &\multirow{4}{4em}{COCO}&Baseline & 40.331 & 39.701 \\
    && DEAL & 40.805 & 40.312 \\
    && ISL/FSL & \textbf{41.719} & \textbf{41.258} \\
    && DEAL + ISL/FSL & 41.583 & 41.186 \\
    \bottomrule
  \end{tabular}
    \label{tab:baseline_results_maxmIoU}
\end{table}
\subsection{DEAL on WeakTr}
With the effect of ISL/FSL verified on WeakTr, we integrate DEAL into the training process, where DEAL is applied twice ---to the edges extracted from the coarse CAM and to the edges extracted from the fine CAM. Similarly to ISL/FSL, we aggregate them into a single score by taking the average, so we optimize
\begin{equation}
    \begin{aligned}
\mathcal{L}_\mathrm{edge} &= \frac{1}{2}(\mathcal{L}_\mathrm{edge,fine} + \mathcal{L}_\mathrm{edge,coarse})\\
\mathcal{L}_\mathrm{total} &= \mathcal{L}_\mathrm{WeakTr} + w_{edge}\mathcal{L}_\mathrm{edge}
    \end{aligned}
    \label{eq:weaktr_loss_deal}
\end{equation}
applying $\mathcal{L}_\mathrm{edge}$ isolated from ISL/FSL. 
We use a weight of $w_{edge} = 0.04$, as we have experienced that higher weights yield collapsed solutions where the CAM only activates the bottom corners for all classes and lower weights didn't yield any effect. Furthermore, since $\mathcal{L}_\mathrm{edge}$ depends on coarsely aligned edges, we keep $w_{edge}  = 0$ for the first 20 epochs. We verify the effect of our approach on both Pascal VOC and MS COCO. The results are listed in Table \ref{tab:baseline_results_maxmIoU}, and we can observe a mean improvement of 1.331 and 1.273 points in mIoU for the train and val split of PASCAL VOC. On MS COCO, applying DEAL leads to an improvement of 0.474 and 0.611 points in mIoU
respectively. The results demonstrate that DEAL can improve segmentation performance across datasets, and generalizes to the validation data too.
\subsection{DEAL on SEAM}
To verify whether DEAL is model-agnostic and improves segmenation performance of other architectures, we repeat the baseline, ISL/FSL and DEAL experiments with another model for CAM generation, i.e. SEAM \cite{wang_self-supervised_2020}. We perform these experiments only on PASCAL VOC, since the focus of this experiment is to verify DEAL's model-agnostic property. Table \ref{tab:seam_baseline_results_maxmIoU}  lists the results of the model trained without any additional loss, with ISL/FSL and with DEAL. We integrate DEAL into SEAM on the model path using the original mini-batch, and apply DEAL twice---to the regular CAM and to the refined CAM using the Pixel Correlation Module of SEAM. Again, we perform a simple averaging, so the total loss optimized is 
\begin{equation}
    \begin{aligned}
    \mathcal{L}_\mathrm{edge} &= \frac{1}{2}\left(\mathcal{L}_\mathrm{edge,cam} + \mathcal{L}_\mathrm{edge, cam_{rv1}}\right)\\
    \mathcal{L}_\mathrm{total} &= \mathcal{L}_\mathrm{SEAM} + w_{edge}\mathcal{L}_\mathrm{edge}
    \end{aligned}
\end{equation}
with $w_{edge} = 0.04$ as with WeakTr. SEAM is trained for a total of 8 epochs, and we keep $w_{edge}=0$ for the first three epochs. Table \ref{tab:seam_baseline_results_maxmIoU} lists the results of this experiment. The positive influence of ISL/FSL on SEAM training is verified. Applying DEAL to SEAM yields a mean improvement of 0.824 and 1.153 points in mIoU for the train and val split of PASCAL VOC respectively. Given that we have not changed the loss weight $w_{deal}$, and see improved segmentation performance on this different, ResNet-based model, the experiments support the model-agnostic property of our proposed DEAL. 
\begin{table}[tbp]
    \centering
    \caption{\normalfont Segmentation performance of CAMs generated by SEAM trained on PASCAL VOC. We report the mIoU of each variant as a mean of four runs with four different seeds.}
    \small
    \setlength{\tabcolsep}{2pt}
    \begin{tabular}{@{}lll>{\centering\arraybackslash}p{0.15\linewidth}>{\centering\arraybackslash}p{0.15\linewidth}@{}}
        \toprule
        \multirow{2}{4em}{\textbf{Model}} & \multirow{2}{4em}{\textbf{Dataset}}& \multirow{2}{4em}{\textbf{Variant}}&
        \multicolumn{2}{c}{\textbf{mIoU} $\uparrow$}  \\
        &&&train&val\\
        \midrule
        \multirow{4}{4em}{\textbf{SEAM}} & \multirow{4}{4em}{VOC}&Baseline& 57.556&54.045 \\
        && DEAL & 58.380 & 55.198\\
        && ISL/FSL & 61.615 & 58.296 \\
        && DEAL + ISL/FSL & \textbf{62.579} & \textbf{59.391} \\
        \bottomrule
  \end{tabular}
    \label{tab:seam_baseline_results_maxmIoU}
\end{table}
\subsection{DEAL on non-standard datasets}
\begin{figure}[t]
    \centering
    \includegraphics[width=0.95\linewidth]{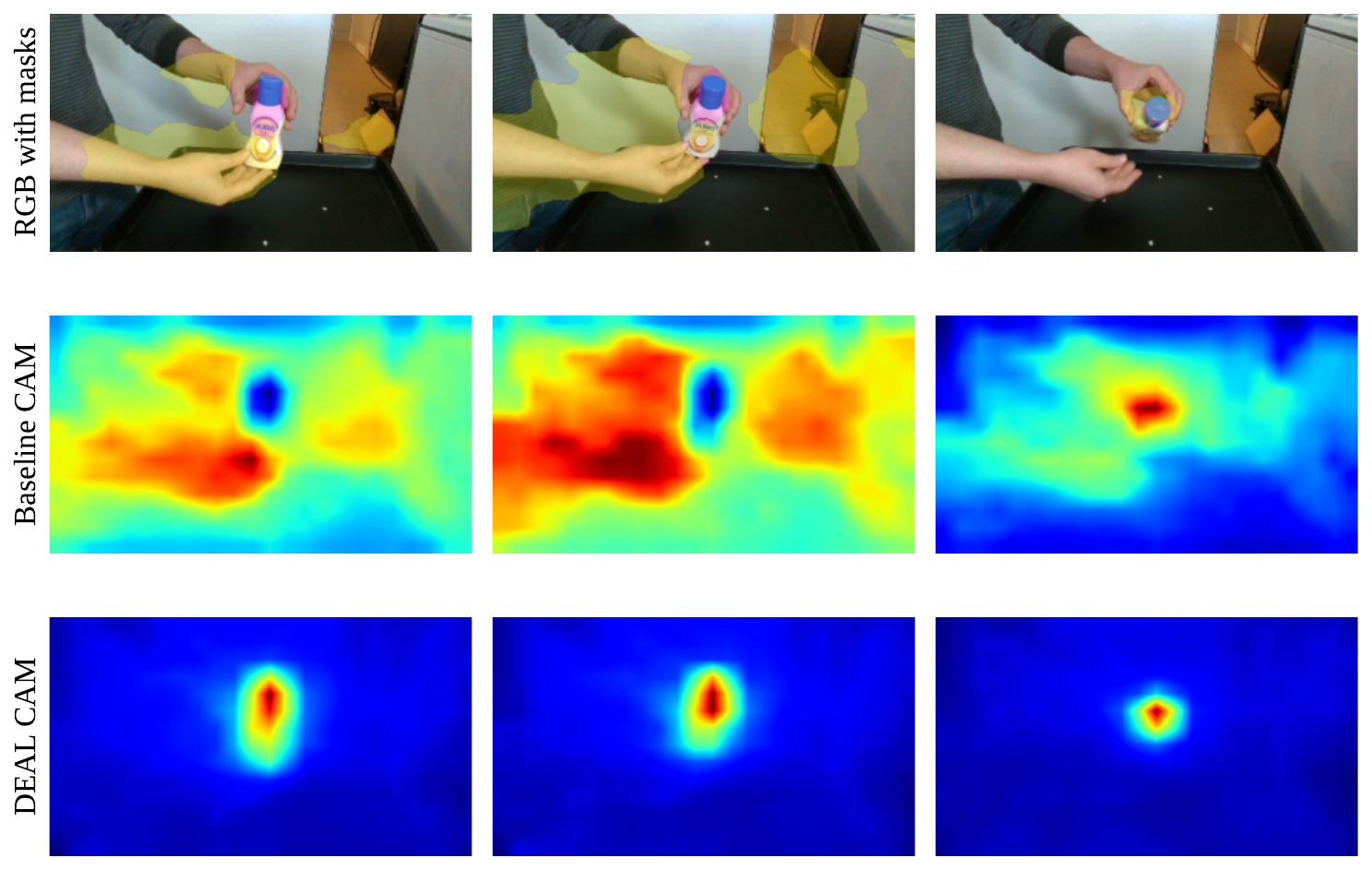}
    \caption{Qualitative results for HOPE, shown on three consecutive frames with a gap of 10 frames in between, from left to right. The top row shows the input RGB images, overlaid with the thresholded CAMs in yellow and magenta for baseline and DEAL respectively. The target object is the bottle being manipulated by the hands. The middle and bottom rows show the CAMs for baseline and DEAL, respectively.}
    \label{fig:bop_qualitative}
\end{figure}

Opposed to other methods that test on COCO and PASCAL only, we choose to additionally verify the beneficial effect of our method on the HOPE dataset \cite{tyree_6-dof_2022}, a pose estimation dataset for robotic manipulation of household objects. In contrast to PASCAL VOC and MS COCO, this dataset contains real RGB-D data from an Intel RealSense camera. We again establish a baseline without any additional losses on WeakTr, and then apply DEAL as outlined in Eq. \ref{eq:weaktr_loss_deal}. Additionally, we want to confirm that our method does not rely on clean, estimated depth maps, so we again generate depth maps with DA-v2b and train WeakTr with DEAL using depth maps from DA-v2b instead of RealSense. We train on the dynamic onboarding sequences and verify our results on the static onboarding sequences. Table \ref{tab:bop_results} lists the results of these experiments, and Fig. \ref{fig:qualitative_results} shows qualitative results.
For the three similar frames in Fig. \ref{fig:bop_qualitative}, DEAL suppresses false positive CAM activations. Additionally, DEAL makes the CAMs be more consistent across object configurations, and yields better segmentation masks, as seen in the magenta overlay in the top row of Fig. \ref{fig:bop_qualitative}.
\begin{table}[tbp]
    \centering
    \caption{\normalfont Segmentation performance of CAMs generated by WeakTr trained on HOPE. We report the mIoU of each variant as a mean of four runs with four different seeds. RealSense and DA-v2b indicate which depth data has been used for training.}
   \small
    \setlength{\tabcolsep}{2pt}
    \begin{tabular}{@{}lll>{\centering\arraybackslash}p{0.15\linewidth}>{\centering\arraybackslash}p{0.15\linewidth}@{}}
        \toprule
        \multirow{2}{4em}{\textbf{Model}} & \multirow{2}{4em}{\textbf{Dataset}}& \multirow{2}{4em}{\textbf{Variant}}&
        \multicolumn{2}{c}{\textbf{mIoU} $\uparrow$}  \\
        &&&dynamic&static\\
        \midrule
        \multirow{3}{4em}{\textbf{WeakTr}} & \multirow{3}{4em}{HOPE}&Baseline& 40.104 & 18.909 \\
        && DEAL (RealSense) & \textbf{51.306} & \textbf{33.715}\\
        && DEAL (DA-v2b)& 50.145 & 32.066 \\
        \bottomrule
  \end{tabular}
    \label{tab:bop_results}
\end{table}
When training WeakTr with DEAL using the RealSense depth maps, we can observe an impressive mean improvement of 11.202 and 14.806 points in mIoU for the static and dynamic sequences of HOPE respectively. When using clean, estimated depth from DA-v2b, this improvement is similar in magnitude, albeit a little less expressed. The results of these experiments show that DEAL can even improve segmentation performance of CAM-based WSSS approaches on non-standard, i.e. robotic datasets like HOPE.  Furthermore, the similar results of real RGB-D vs. estimated RGB-D data demonstrates that DEAL is robust to noise in depth data, making it applicable to real-life robotic datasets.
\subsection{Ablation Studies}
\begin{figure*}[t]
    \centering
    \begin{tabular}{
    >{\centering\arraybackslash}p{0.131\linewidth}
    >{\centering\arraybackslash}p{0.12\linewidth}
    >{\centering\arraybackslash}p{0.12\linewidth}
    >{\centering\arraybackslash}p{0.12\linewidth}
    >{\centering\arraybackslash}p{0.12\linewidth}
    >{\centering\arraybackslash}p{0.12\linewidth}
    >{\centering\arraybackslash}p{0.07\linewidth}}
    Image & GT & Baseline & DEAL & ISL/FSL & DEAL + ISL/FSL  \\
    \end{tabular}
    \includegraphics[width=0.85\linewidth]{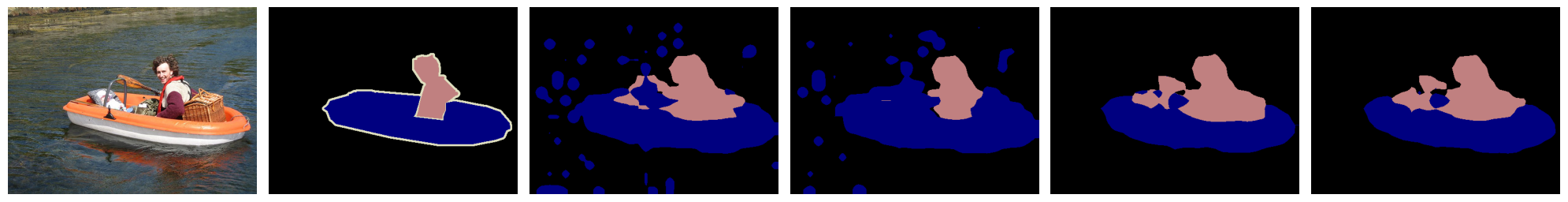}
    \includegraphics[width=0.85\linewidth]{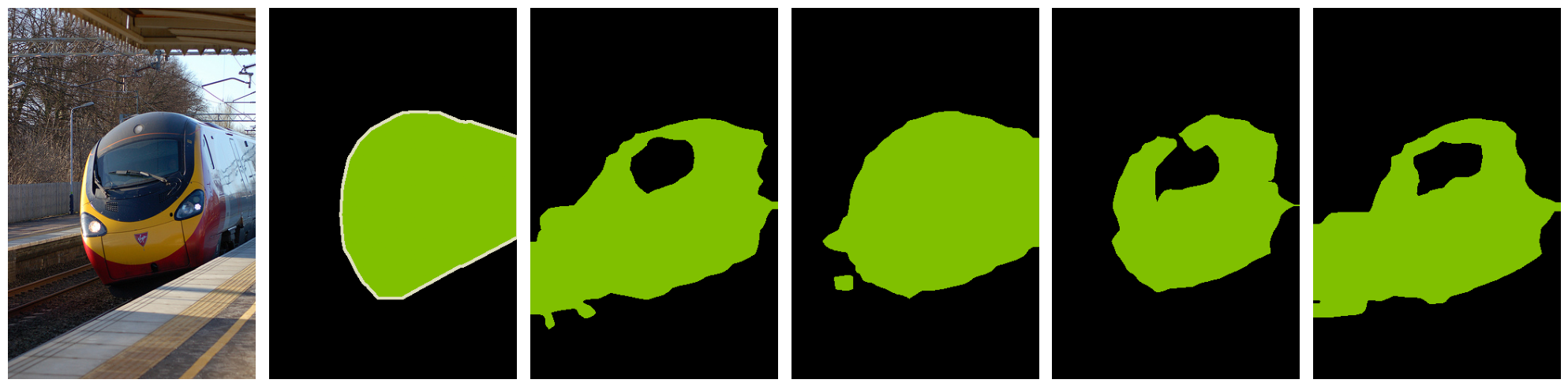}
    \includegraphics[width=0.85\linewidth]{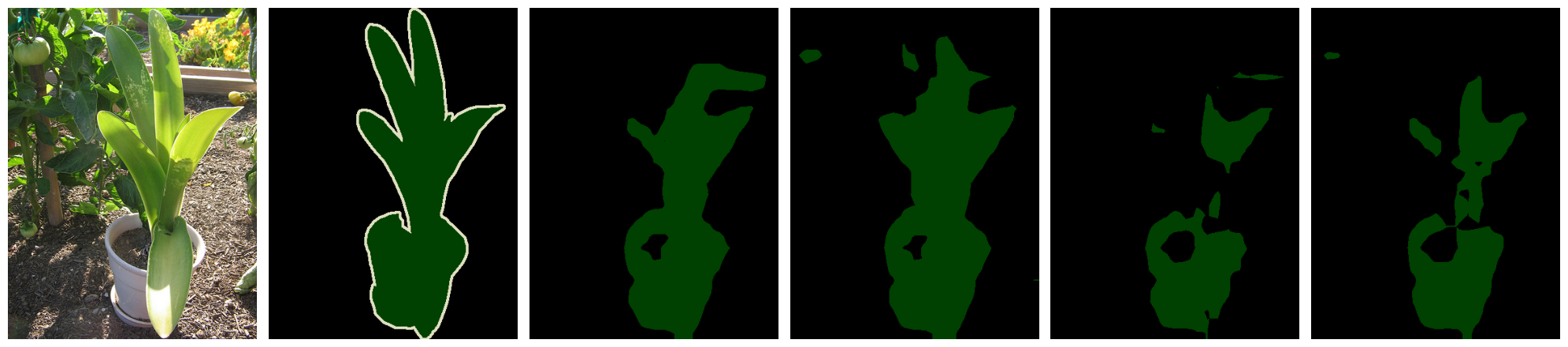}
    \includegraphics[width=0.85\linewidth]{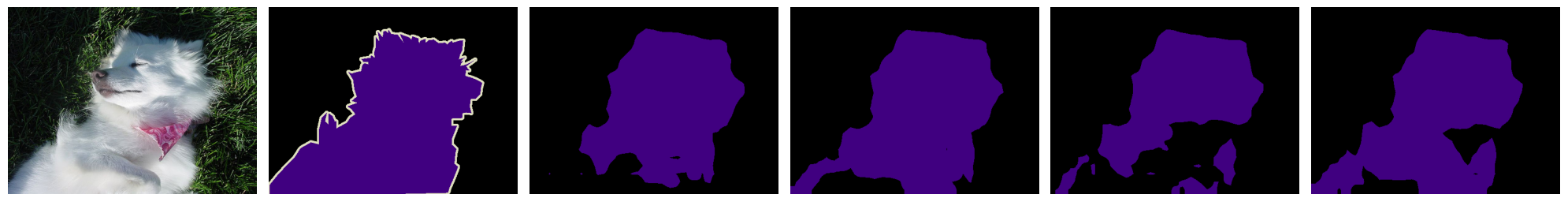}
    \includegraphics[width=0.85\linewidth]{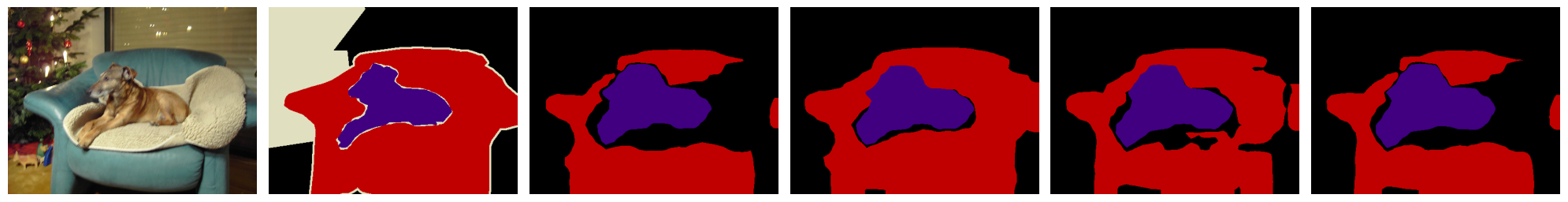}
     \includegraphics[width=0.85\linewidth]{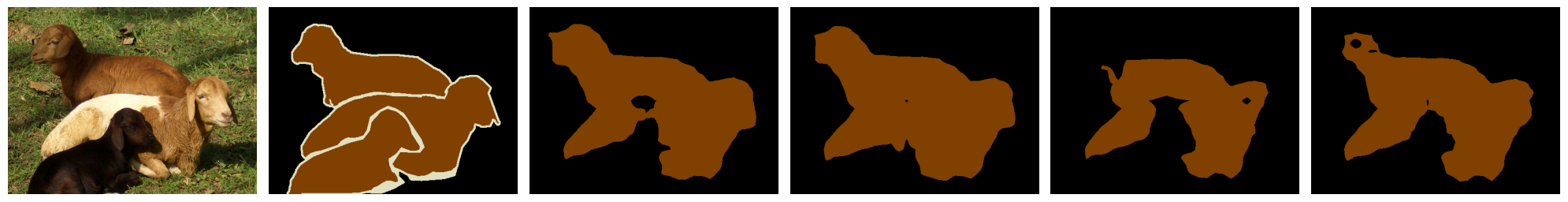}
    \caption{Qualitative results of WeakTr trained with the different variants presented in Table \ref{tab:baseline_results_maxmIoU}. Note that the predicted masks are obtained by CAM thresholding without any post-processing.}
    \label{fig:qualitative_results}
\end{figure*}
Finally, we want to analyze whether the beneficial effect of DEAL can be combined with auxiliary losses to combine the beneficial effects of each loss and yield even better segmentation performance. We combine DEAL with ISL/FSL, train WeakTr on PASCAL VOC and MS COCO, train SEAM on PASCAL VOC and summarize the results in an ablation table in Table \ref{tab:ablation_studies}. Note that, in line with the common practice, the reported values are obtained by taking the best run across four different seeds for each variant. In contrast to this, Tables \ref{tab:baseline_results_maxmIoU} and \ref{tab:seam_baseline_results_maxmIoU} listed the mean mIoU score across four seeds for the trainings with DEAL + ISL/FSL, demonstrating a mean improvement as well. We observe that ISL/FSL can improve the performance of DEAL on PASCAL VOC even further, yielding an mIoU of the generated CAMs of 66.876 and 64.671 for the train and validation splits respectively. This is an improvement of +0.561 / +0.612 points compared to WeakTR + ISL/FSL and +3.612 / +4.432 points compared to the baseline for the train and validation splits respectively. On SEAM, adding DEAL brings an improvement of +0.946 / +1.442 mIoU points on PASCAL VOC train/validation, compared to the baseline. Adding ISL/FSL to DEAL, the overall improvement compared to the baseline is +5.132 and +5.439 points in mIoU for PASCAL VOC train and validation respectively. On HOPE, adding DEAL leads to an impressive improvement of +12.368 points compared to the baseline training with WeakTr. Finally, Fig. \ref{fig:qualitative_results} shows qualitative results of this ablation study on samples from PASCAL VOC. 
\begin{table}[!t]
    \centering
\caption{\raggedright\normalfont Results of the ablation study. We pick the best-performing seed of each variant and compare those.}
\label{tab:ablation_studies}
    \small
    \setlength{\tabcolsep}{2pt}
    \begin{tabular}{@{}ll
    >{\centering\arraybackslash}p{0.15\linewidth}
    >{\centering\arraybackslash}p{0.15\linewidth}    >{\centering\arraybackslash}p{0.15\linewidth}
    >{\centering\arraybackslash}p{0.15\linewidth}
    @{}}
    \toprule
    \textbf{Model} & \textbf{Dataset} & \textbf{DEAL} & \textbf{ISL/FSL} & $\Delta_{\mathrm{train}}\uparrow$ & $\Delta_{\mathrm{val}}\uparrow$\\
    \midrule
    \multirow{10}{0.15\linewidth}{\textbf{WeakTr}}
    &\multirow{4}{0.1\linewidth}{VOC} & & & \textit{64.205} & \textit{61.125} \\
    && \checkmark & & +0.942 & +1.261 \\
    && & \checkmark & +3.051 & +3.820 \\
    && \checkmark & \checkmark & \textbf{+3.612} & \textbf{+4.432}\\
    \cmidrule(lr){2-6}
    &\multirow{4}{0.1\linewidth}{COCO} & & & \textit{40.918} & \textit{40.260} \\
    && \checkmark & & +0.358 & +0.529 \\
    && & \checkmark & \textbf{+1.374} & \textbf{+1.586} \\
    && \checkmark & \checkmark & +0.990 & +1.274 \\
    \cmidrule(lr){2-6}
    &\multirow{2}{0.1\linewidth}{HOPE} & & & \textit{51.136} & \textit{23.998}\\
    &&\checkmark&&\textbf{+12.368}&\textbf{+16.416}\\
    \midrule
    \multirow{4}{0.15\linewidth}{\textbf{SEAM}} & \multirow{4}{0.1\linewidth}{VOC} & & &\textit{57.618}&\textit{54.139}\\
    &&\checkmark&&+0.946&+1.442\\
    &&&\checkmark&+4.144&+4.298\\
    &&\checkmark&\checkmark&\textbf{+5.132}&\textbf{+5.439}\\
    \bottomrule
  \end{tabular}
\end{table}
\section{Conclusion}
This study presented our work on improving the segmentation performance of CAM-based WSSS by integrating depth information through our Depth Edge Alignment Loss (DEAL). Our approach adds a loss to the training of CAM-based WSSS models and intuitively encourages alignment of CAM edges and depth edges, exploiting the RGB-D modality widely available in robotics. 
We have demonstrated how DEAL improves mIoU performance across datasets, have shown that DEAL is model-agnostic by applying it to two different WSSS frameworks using different vision architectures, and finally have shown that even noisy, real-life depth data from a robotic dataset can improve mIoU performance in a WSSS setting. We see great potential of WSSS in a robotic context, given data-related challenges that roboticists face when applying systems to new domains. With the current trend of WSSS frameworks using Vision-Language models, we intend to analyze how e.g. prompting those for depth estimation can be combined with ground-truth depth information and how this could improve common feature representations for WSSS. 
\bibliographystyle{IEEEtran}
\bibliography{IEEEabrv,references_new_no_url}
\end{document}